\documentclass[conference]{IEEEtran}
\IEEEoverridecommandlockouts

\usepackage{cite}
\usepackage{amsmath,amssymb,amsfonts}
\usepackage{algorithmic}
\usepackage{graphicx}
\usepackage{textcomp}
\usepackage{xcolor}
\usepackage{hyperref}
\usepackage{geometry}
\def\BibTeX{{\rm B\kern-.05em{\sc i\kern-.025em b}\kern-.08em
    T\kern-.1667em\lower.7ex\hbox{E}\kern-.125emX}}
\begin{document}

\title{Vision-based Perception System for Automated Delivery Robot–Pedestrians Interactions\\
\thanks{This research is funded by Canada Research Chair in Disruptive Transportation Technologies and Services (CRC-2022-00480).}
}

\author{\IEEEauthorblockN{Ergi Tushe and Bilal Farooq}
\IEEEauthorblockA{\textit{Laboratory of Innovations in Transportation (LiTrans) and Data Science Program} \\
\textit{Toronto Metropolitan University}, 
\textit{Canada, Toronto} \\
etushe@torontomu.ca and bilal.farooq@torontomu.ca}\\
\textbf{Cite as:}\\
\textbf{Tushe E., Farooq, B., (2025) Vision-based perception system for automated delivery robot-pedestrian interactions,}\\
\textbf{In the proceedings of 11th IEEE International Smart Cities Conference (ISC2), Patras, Greece, October 2025.}}

\maketitle

\begin{abstract}
The integration of Automated Delivery Robots (ADRs) into pedestrian-heavy urban spaces introduces unique challenges in terms of safe, efficient, and socially acceptable navigation. We develop the complete pipeline for a single vision sensor based multi-pedestrian detection and tracking, pose estimation, and monocular depth perception. Leveraging the real-world MOT17 dataset sequences, this study demonstrates how integrating human-pose estimation and depth cues enhances pedestrian trajectory prediction and identity maintenance, even under occlusions and dense crowds. Results show measurable improvements, including up to a 10\% increase in identity preservation (IDF1), a 7\% improvement in multiobject tracking accuracy (MOTA), and consistently high detection precision exceeding 85\%, even in challenging scenarios. Notably, the system identifies vulnerable pedestrian groups supporting more socially aware and inclusive robot behaviour.
\end{abstract}


\section{Introduction}\label{intro}
Automated Delivery Robots (ADRs) have emerged as a promising alternative for urban logistics, navigating sidewalks and shared spaces while interacting with pedestrians. ADRs are particularly well-suited for urban settings where high population density, short delivery distances, and growing demands for sustainable transport solutions challenge conventional logistics methods. The expansion of ADR deployment aligns with global trends prioritizing sustainability and efficiency. For example, the growth of e-commerce has intensified the need for scalable last-mile delivery solutions. Between 2005 and 2019, e-commerce in Sweden grew by 18\% annually, followed by a sharp 40\% rise during the COVID-19 pandemic \cite{alverhed2024autonomous}. The surge in e-commerce demand has strained conventional delivery infrastructure, making ADRs a practical solution to alleviate congestion, reduce carbon emissions, and meet delivery efficiency requirements. 

However, the successful integration of ADRs into urban environments is fraught with challenges stemming from the unpredictable nature of pedestrian behaviour. Unlike automated vehicles, which typically operate on structured roads with defined lanes and signals, ADRs must navigate dynamic and unstructured environments where social norms and human movements dictate safe and acceptable navigation. Pedestrian pathways are inherently more unpredictable than vehicular roads, as they involve varied pedestrian types, diverse walking patterns, and spontaneous human behaviour and interactions. ADRs must be capable of interpreting complex human behaviour, predicting pedestrian motion, and making real-time navigation decisions that prioritize safety and efficiency. Human-pose estimation plays a crucial role in enhancing ADR perception. By analyzing the relative location, direction, human-joint positions, and body postures, ADRs can infer pedestrian intent and adjust navigation strategies accordingly. This is particularly vital when navigating around vulnerable pedestrian groups such as individuals using mobility aids, children, or distracted walkers \cite{taylor2020robot}. Furthermore, ADRs must recognize pedestrian group formations, handle interactions with people who exhibit erratic or unexpected movements and differentiate between individuals moving alone versus groups moving in unison. The integration of stereo cameras or LiDAR sensors could significantly enhance ADR's perception system. However, such solutions greatly increase computational costs and power consumption, making them less feasible for lightweight ADR deployments at scale.

Using MOT17 dataset, this research directly contributes to enhancing ADR perception and navigation capabilities within pedestrian-rich urban environments by developing a single vision sensor based perception system for ADRs that integrates pedestrian detection, tracking, pose estimation, and monocular depth estimation. The focus is on covering the pedestrian heterogeneity in urban scenes. Thus enabling the socially aware ADR navigation in real-world urban environments, offering a scalable, efficient alternative to multi-sensor setups while advancing human-robot interaction and adaptability. To ensure quick deployment, generalizability, and fast-processing, the perception system is developed using pretrained models, such as You Only Look Once (YOLO) \cite{wang2024yolov9}, Simple Online and Realtime Tracking with a Deep Association Metric (DeepSort) \cite{belmouhcine2021robust}, and Depth-Anything \cite{depthanything}, as the core elements.

\section{Background}\label{background}

A key complexity in urban ADR deployment is the need for socially aware navigation, particularly in contexts involving vulnerable users. Pedestrians with mobility aids, distracted pedestrians, or groups of children present additional challenges for ADRs, necessitating adaptable perception and planning capabilities. ADRs must identify diverse pedestrian characteristics and predict pedestrian intent to ensure safe and socially compliant navigation. 

Previous research has explored ADR applications in specialized contexts to address pedestrian-related challenges. For instance, Hanheide et al. \cite{hanheide2017and} demonstrated the potential for adaptive scheduling and spatio-temporal modelling in a care home environment, improving the relevance of information delivery. Similarly, ADRs have been employed as assistive technologies for visually impaired individuals, using tactile feedback and accurate localization to support navigation in complex spaces \cite{goodrich2008human}. In educational and therapeutic settings, ADRs equipped with emotive capabilities have supported children with autism spectrum disorders, facilitating structured social interactions. In retail contexts, culturally aware ADRs have enhanced customer engagement by adapting communication strategies to suit diverse social groups. However, limitations persist in current ADR perception systems. Many rely on single-frame pedestrian detection, which struggles with maintaining identity consistency across multiple frames. Challenges such as occlusions, rapid pedestrian movement, and the unpredictable nature of human behaviour further complicate ADR navigation \cite{hwang2024emma}. Pedestrian detection models often fail to maintain identity tracking during instances of pedestrian overlap, leading to degraded tracking quality and increased risks of unsafe or socially inappropriate navigation.

Recent advances in deep learning have enabled the development of robust pedestrian detection models and human-pose estimation algorithms \cite{taylor2020robot}. These advancements serve as foundational components for developing ADRs capable of more stable and socially aware navigation. However, their integration into ADR systems operating under resource constraints and real-time requirements remains underexplored. One fundamental challenge in pedestrian detection and tracking for ADRs is differentiating between movement caused by the robot's own motion and actual pedestrian motion. When cameras are mounted on moving ADRs, sensor motion can be misinterpreted as pedestrian movement, leading to false detections or the failure to recognize stationary pedestrians. This issue is particularly problematic when ADRs navigate uneven terrain or experience sudden orientation changes. Advanced tracking models such as DeepSORT and Observation-Centric SORT (OC-SORT) attempt to mitigate this problem by employing Kalman filters, which predict the future positions of detected pedestrians based on prior motion patterns to reduce misclassifications \cite{wang2024yolov9}. However, these methods still require precise calibration to account for motion-induced distortions in the ADR's field of view. Xiao and Feng \cite{xiao2023multi} introduced an improved bounding box regression method using Generalized Intersection over Union (GIOU), which enhances bounding box consistency across frames, reducing inaccuracies caused by camera shifts. Despite these refinements, distinguishing pedestrian motion from background movement remains a persistent challenge, particularly in highly dynamic environments.

Another major limitation stems from the differences between exocentric and egocentric vision in pedestrian detection. Traditional pedestrian detection models rely on exocentric vision, where cameras are positioned externally, such as on traffic poles or infrastructure-mounted sensors. While effective in static monitoring scenarios, exocentric vision lacks adaptability for real-time ADR navigation, as it does not capture the changing perspective of a moving automated robot. Conversely, egocentric vision, where cameras are mounted directly on the ADR, provides a continuous real-time view of the environment but introduces new challenges in detecting pedestrians within a constantly shifting frame. To improve egocentric vision-based detection, Xiao and Feng \cite{xiao2023multi} integrated a YOLO-based model with enhancements such as Soft Non-Maximum Suppression (SoftNMS) and GhostNet, improving object recognition even in crowded settings. However, achieving robust pedestrian detection in unstructured environments requires further advancements in feature extraction and motion compensation techniques, as occlusions and erratic pedestrian movement continue to degrade model performance.

Depth ambiguity remains one of the most persistent challenges in pedestrian detection, as 2D images fail to provide reliable distance estimations. In dense urban environments, pedestrians may assume various poses, be partially occluded, or blend into complex backgrounds, making it difficult for ADRs to accurately determine proximity and movement intent. YOLO-based detection models attempt to address this issue using modules such as Spatial Pyramid Pooling Fusion (SPPF), which aggregates contextual information across multiple spatial scales to improve detection robustness \cite{xiao2023multi}. While these methods improve size and distance perception, they do not fully resolve depth-related ambiguities, especially in poor lighting or high-traffic areas. The integration of stereo cameras or LiDAR sensors could significantly enhance depth estimation; however, such solutions increase computational costs and power consumption, making them less feasible for lightweight ADR deployments.

Furthermore, unpredictable pedestrian movements are a significant challenge in accurate individual detection and tracking. Unlike vehicles, which follow predefined traffic rules and exhibit relatively structured motion, pedestrian behaviour is highly erratic, often involving sudden stops, rapid direction changes, or group interactions. Traditional tracking models rely on motion prediction algorithms to anticipate pedestrian movement, but behaviour modelling remains an area of active research. Current tracking frameworks, such as appearance-based re-identification and motion prediction networks, attempt to correlate pedestrian behaviour across frames, but they still face limitations in fully capturing pedestrian intent. For example, while trajectory-based models can estimate motion patterns, they struggle to anticipate interactions such as a pedestrian suddenly changing direction to avoid an obstacle or engaging in social interactions that alter their trajectory. As pedestrian tracking shifts from simple trajectory prediction to behaviour modelling, integrating human pose estimation will be essential for achieving a more robust understanding of pedestrian interactions.

\begin{figure}[tbp]
\centerline{\includegraphics[scale=0.45]{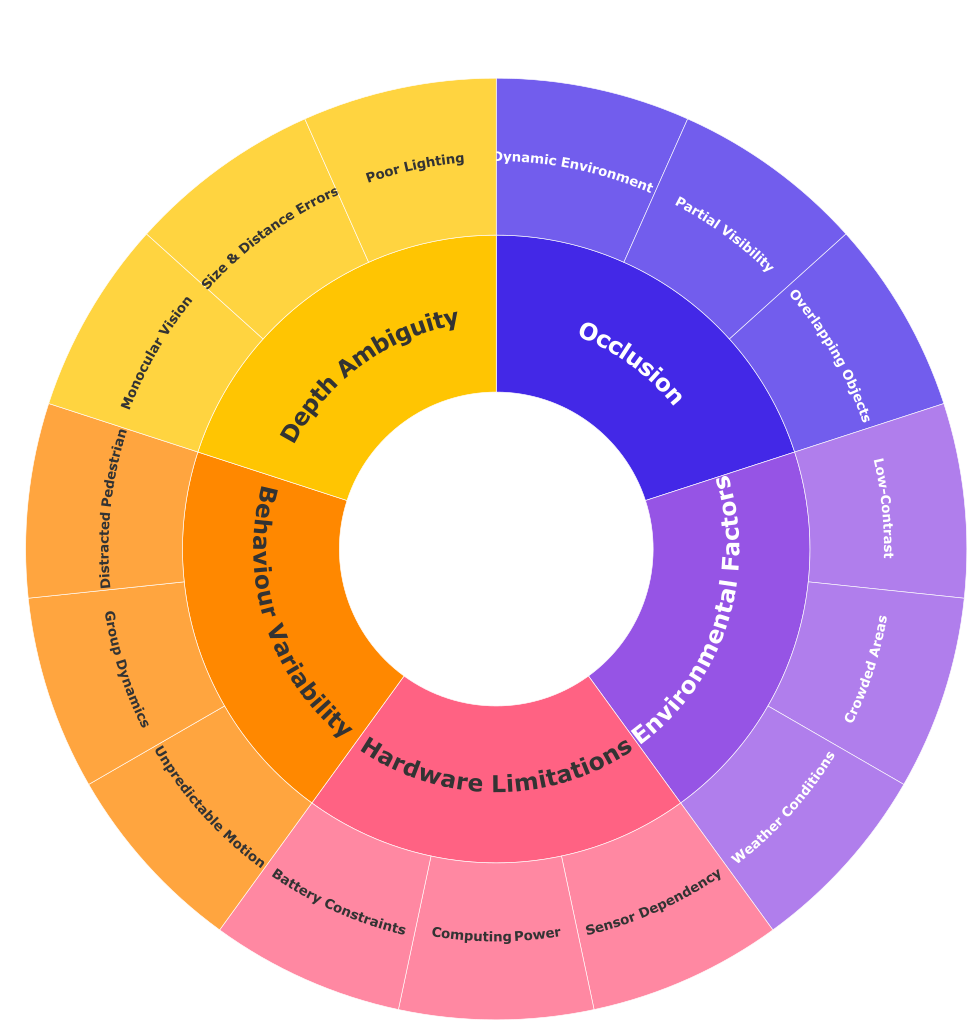}}
\caption{Overall challenges of pedestrian-ADR interactions}
\label{adr_lim}
\end{figure}

Occlusion remains one of the most disruptive factors in pedestrian tracking, particularly in dense urban settings where vehicles, buildings, or other pedestrians frequently block an ADR's line of sight. Tracking failures occur when a pedestrian disappears for an extended period, leading to identity switches or lost trajectories. To mitigate this, YOLO-based detection models are increasingly paired with tracking enhancements such as Soft Non-Maximum Suppression (SoftNMS), which assigns weights to overlapping bounding boxes instead of outright suppressing them, reducing missed detections \cite{xiao2023multi}. Re-Identification (Re-ID) networks have also been introduced to maintain pedestrian identities by leveraging appearance-based feature matching across frames, ensuring that pedestrians can be re-associated after occlusions. Multi-frame association models further improve tracking performance by predicting pedestrian movement across frames to maintain continuity in identity tracking. While these techniques enhance short-term tracking accuracy, prolonged occlusions still result in frequent identity mismatches, highlighting the need for more robust scene understanding and long-term pedestrian re-identification strategies. Fig. \ref{adr_lim} highlights the overall challenges in terms of the research on pedestrian-ARD interactions.

\section{Methodology}
\begin{figure*}[htbp]
\centerline{\includegraphics[scale=0.58]{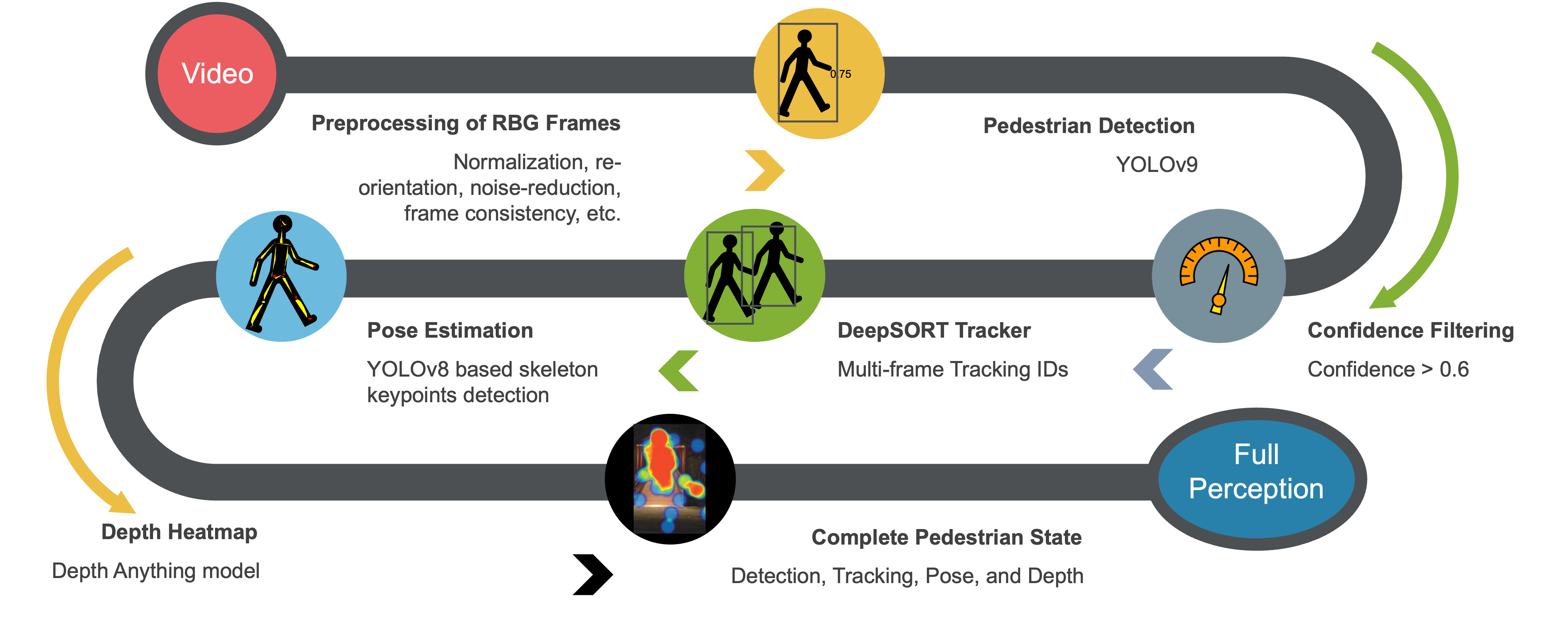}}
\caption{Workflow of ADR perception pipeline}
\label{wf1}
\end{figure*}

The perception pipeline is designed using pre-trained real-time models to provide the ADR with a comprehensive understanding of pedestrian movement within complex urban environments. The system operates in four major stages (Fig. \ref{wf1}): 
\begin{enumerate}
    \item Pedestrian Detection using YOLOv9,
    \item Pedestrian Tracking using DeepSORT,
    \item Human Pose Estimation using YOLO-Pose, and
    \item Monocular depth estimation using Depth-Anything.
\end{enumerate}
Each component addresses specific challenges encountered in dynamic and crowded urban scenes, with a particular emphasis on detecting diverse pedestrian profiles, including vulnerable groups such as the elderly, children, and individuals using mobility aids, who often require additional spatial consideration in navigation systems. The use of pre-trained models also ensures the generalizability, scalability, robustness, and real-time deployment of the perception system. 

The preprocessing of MOT17 dataset involved the application of contrast adjustments, brightness normalization, resizing, re-orientation, noise reduction and frame consistency checks. Sequences containing heavy occlusion events were tagged separately, allowing for more focused evaluation of the model’s ability to handle identity switches and pedestrian re-identification. Given the real-time processing constraints of ADRs operating in complex urban environments, this study employs YOLO v9 for pedestrian detection, leveraging its speed and efficiency in handling dense scenes with over-crowded pedestrians. Unlike two-stage object detectors, which separately propose candidate regions before classification, YOLO is a single-stage detector, allowing it to perform both tasks simultaneously, thus ensuring real-time performance without sacrificing accuracy \cite{ding2024research}. This step results in the model performing inference on each frame, predicting bounding box coordinates, class labels, and confidence scores for detected pedestrians. After developing a sensitivity analysis, a confidence threshold of 0.6 is applied to ensure that only high-probability detections are retained. This step significantly reduces false positives caused by ambiguous background elements, shadows, or overlapping pedestrians. 

YOLO, by design, struggles with partial occlusions, as standard detection models often fail to recognize pedestrians when a significant portion of their body is obstructed. To mitigate this, Soft Non-Maximum Suppression (SoftNMS) is integrated, which assigns weighted penalties to overlapping bounding boxes instead of outright discarding them. This adjustment allows for the retention of partially visible pedestrians, reducing the rate of missed detections \cite{sun2020survey}. Additionally, YOLO-based detection faces challenges when camera motion introduces motion blur or illumination conditions change dynamically. To account for this, adaptive confidence thresholding is implemented, dynamically adjusting the detection sensitivity based on environmental factors. Temporal smoothing techniques are applied, ensuring that YOLO's detections remain stable across consecutive frames. Additionally, multi-frame association methods enable identity continuity even when pedestrians momentarily disappear behind obstacles.

DeepSORT is integrated into the detection pipeline, allowing for long-term identity preservation \cite{belmouhcine2021robust}. DeepSORT utilizes a Kalman filter-based motion prediction model, which estimates the future position of each detected pedestrian based on their previous movements. This predictive capability is particularly beneficial in scenarios involving occlusions, where a pedestrian may temporarily disappear behind an object or another individual. By leveraging historical tracking data, DeepSORT can anticipate pedestrian reappearance, thereby reducing identity switches and enhancing tracking stability. After a sensitivity analysis, the max\_age parameter was set to 50, allowing pedestrian identities to persist for up to 50 frames before being reassigned, ensuring continuity even if a pedestrian momentarily disappears from the frame. The nn\_budget parameter, set at 150, expands the model’s capacity to store appearance-based feature embeddings, improving its ability to re-identify pedestrians despite occlusions or changes in pose. Additionally, the max\_iou\_distance parameter was set to 0.7, fine-tuning the intersection-over-union (IoU) threshold to ensure that pedestrian tracking remains reliable even in crowded urban environments where bounding boxes may overlap. 

YOLOv8-Pose, which has a built-in pose estimation model, was customized to extract skeletal representations of pedestrians from the MOT17 dataset. The dataset's diverse conditions, including variations in lighting, crowd density, and occlusions, provide a challenging test environment for pose estimation models. The implementation involved processing each image sequence through the YOLOv8-Pose model, which predicted key points corresponding to 17 major human joints. These key points were then mapped onto the original image sequences, generating skeletal representations that visually depict the specific pedestrian posture and motion. The use of pre-trained YOLOv8-Pose weights, fine-tuned on human pose datasets, enabled the model to generalize well across different pedestrian poses and movements within the MOT17 sequences.

A learning-based approach using the Depth-Anything model \cite{depthanything} is used to develop the depth estimation. This model generates dense pixel-wise depth heat maps, producing relative depth estimates for each pixel in the scene. Unlike heuristic methods, Depth-Anything leverages large-scale datasets to learn depth cues implicitly, including shading, texture gradients, perspective, and object occlusion patterns. The resulting heat maps provide a more informative third dimension, which not only improves detection and tracking but also contextualizes the scene.
\begin{figure*}[]
\centerline{\includegraphics[scale=0.825]{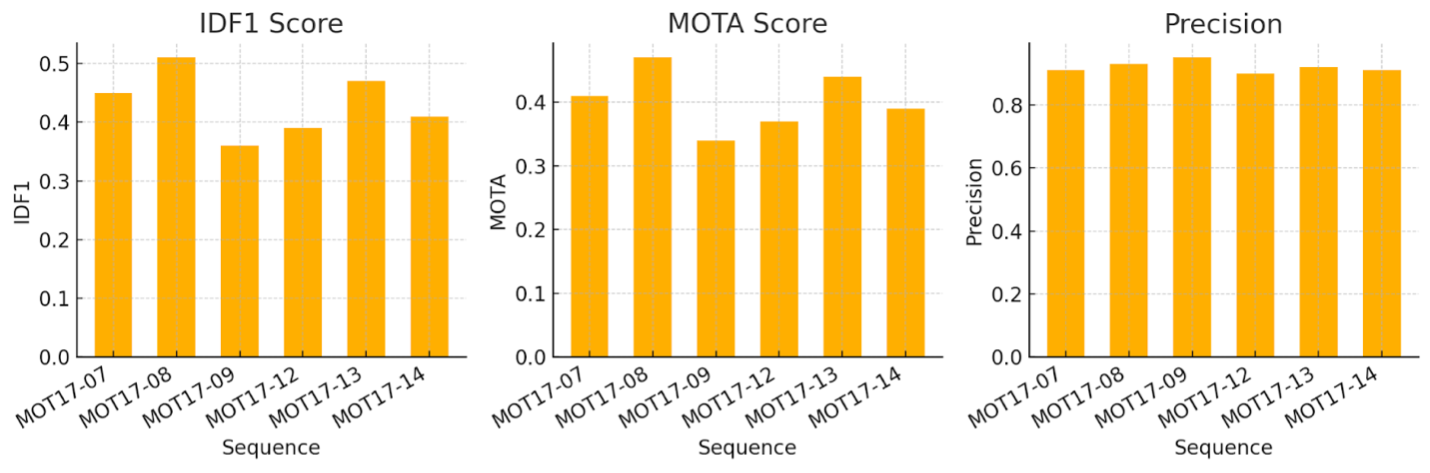}}
\caption{ System performance on MOT17 sequences using IDF1, MOTA, and Precision
}
\label{qres}
\end{figure*}
\begin{figure*}[]
\centerline{\includegraphics[scale=0.3275]{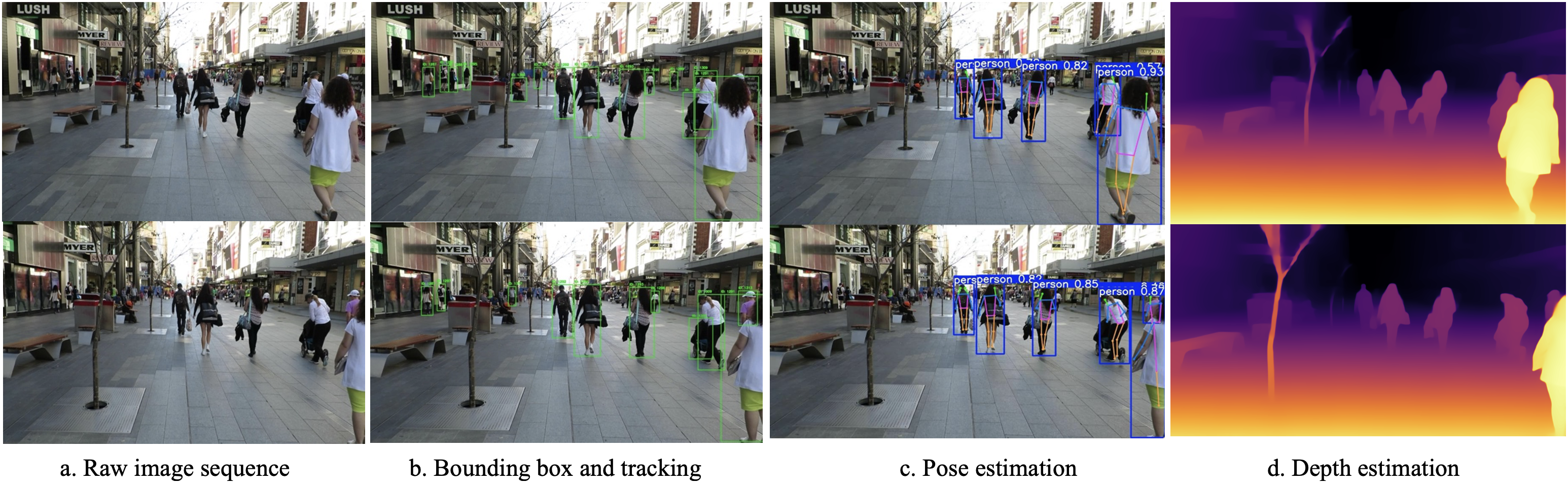}}
\caption{Examples of the perception pipeline, including detection, tracking, pose estimation, and depth estimation over multiple frames. For complete videos, please refer to: \url{https://www.youtube.com/playlist?list=PL5asEN_LoEU2KerAnzwkazFhNnL9j3-wz}.}
\label{vres}
\end{figure*}

\section{Results}
The results obtained in this study confirm that integrating depth and pose estimation into a pedestrian detection and tracking pipeline significantly enhances system performance under a variety of urban conditions. Across the MOT17 dataset (Fig. \ref{qres}), the system achieved Identity F1 (IDF1) scores ranging from 0.36 to 0.52 and Multi-Object Tracking Accuracy (MOTA) values from 0.34 to 0.47. In comparison, the baseline YOLO + DeepSORT systems in prior literature \cite{razzok2023pedestrian, zheng2023deep} typically achieved IDF1 values between 0.34 and 0.42 and MOTA values below 0.40. This translates into an approximate 8–10\% improvement in identity preservation and an increase of up to 7\% in tracking accuracy. These improvements are particularly prominent in sequences with moderate pedestrian densities and occlusions, such as MOT17-08 and MOT17-13, where IDF1 scores exceeded 0.50. In these cases, the additional spatial information provided by monocular depth maps and the enhanced motion cues from pose estimation substantially reduced identity switches and track fragmentation. Notably, precision remained consistently high across all sequences (above 0.85), with peak values nearing 0.95, outperforming typical baseline results without compromising real-time operation.

In terms of visual output, Fig. \ref{vres} shows an example of the complete pipeline. The perception system is able to detect pedestrians with high accuracy, despite them being at varying distances in the scene. It is also able to track individuals over multiple frames despite being partially occluded. Furthermore, the system is able to estimate a range of complex human poses (e.g., walking while carrying objects or bending to check the kid in the stroller) as well as the relative distances from static and dynamic objects accurately. To experience the performance of the system on longer sequences, we refer the readers to watch the videos provided at the link provided in the caption of Fig. \ref{vres}.

\begin{figure}[]
\centerline{\includegraphics[scale=0.35]{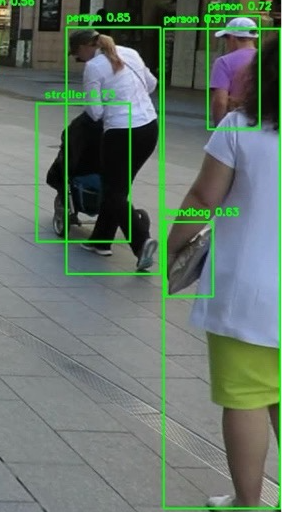}\space \space \space \includegraphics[scale=0.41]{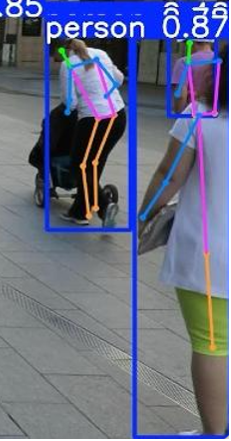}}
\caption{Detection and labelling of the stroller in the scene}
\label{vresz}
\end{figure}

Fig. \ref{vresz} shows further details in the scene, particularly the detection and labelling of a pedestrian and the stroller they are pushing. Similarly, the system is also able to detect the other pedestrian and the handbag they are carrying. The system's ability to detect and maintain consistent identities for vulnerable pedestrian groups, such as the person pushing a stroller with a kid, demonstrates the practical implications of this pipeline for real-world ADR applications. Vulnerable pedestrians often move at slower speeds, occupy larger spatial footprints, or make abrupt movements, which can complicate perception and prediction tasks. The successful tracking of such individuals across multiple frames underscores the pipeline's capacity for socially responsive navigation. This feature is especially significant in pedestrian-rich urban environments, where socially aware behaviour enhances not only safety but also public acceptance. By identifying auxiliary mobility objects (e.g., strollers), the system can contribute to more inclusive robotic interactions, reinforcing trust and supporting broader adoption of ADR technologies in diverse communities. 

\section{Conclusions}
The results demonstrate that the proposed system successfully detects, tracks, and interprets pedestrian movements in complex urban environments. Both qualitative outputs and quantitative metrics highlight the system’s ability to operate under varied conditions, with consistent effectiveness in moderately crowded scenes. Of particular importance was the system’s ability to detect and consistently track vulnerable pedestrians, such as a woman pushing a stroller, a real-world scenario that underscores the significance of socially aware navigation in ADR deployment. Such capacity directly addresses public concerns regarding safety and empathy in shared urban spaces, as successful ADR integration will depend not only on performance metrics but also on social trust and acceptance. The inclusion of pose and depth information enables richer behavioural interpretation, such as walking speed, gait, or proximity, that allows ADRs to make navigation decisions that respect social norms and the needs of diverse pedestrian groups. In this way, the system marks a meaningful step forward in aligning ADR perception with public expectations and ethical standards for urban deployment.

A critical future extension involves advancing the system's capacity for understanding pedestrian intent. Pedestrian kinematics, including velocity, acceleration, and subtle motion cues such as hesitation, deceleration, or lateral movement, provide valuable information about pedestrian intentions and likely future trajectories. Integrating these features into prediction models could enhance ADR's ability to proactively adjust navigation strategies to anticipate pedestrian actions. Temporal modelling techniques, such as Transformer-based architectures, could be employed to capture these spatio-temporal dependencies effectively. Beyond perception and prediction, future work should also focus on improving ADR behaviour through socially aware planning \cite{vasquez2019multi}. ADRs must not only avoid collisions, but also navigate in ways that align with human expectations and social norms. Incorporating kinematic cues and pedestrian intent prediction into behaviour planning would allow ADRs to better balance efficiency with comfort and trustworthiness during pedestrian interactions. Finally, to ensure practical readiness, the system should be further validated on ADR-specific datasets and simulation environments featuring realistic pedestrian dynamics, varying densities, and complex interaction scenarios. Virtual reality based simulation platforms \cite{farooq2018virtual} combining ADR control with pedestrian virtual environments will facilitate large-scale testing under diverse and controlled conditions before real-world deployment. Such simulations will also enable experimentation with interaction strategies and the evaluation of ADR behaviour beyond safety metrics, including trust, acceptance, and social compatibility.

\bibliographystyle{ieeetr}
\bibliography{IEEEfull}

\end{document}